\documentclass[review]{elsarticle}

\usepackage{amsmath}
\usepackage{relsize}
\usepackage{lineno,hyperref}
\usepackage{relsize}
\usepackage{tikz}
\usepackage{bm}
\usepackage{graphicx,float}
\usepackage{xurl}
\modulolinenumbers[5]

\journal{Neurocomputing}

\newcommand{\xoverbrace}[2][\vphantom{\dfrac{A}{A}}]{\overbrace{#1#2}}

\newcommand*\circled[1]{\tikz[baseline=(char.base)]{
            \node[shape=circle,draw,inner sep=2pt] (char) {#1};}}

\makeatletter
\newcommand*{\DivideLengths}[2]{%
  \strip@pt\dimexpr\number\numexpr\number\dimexpr#1\relax*65536/\number\dimexpr#2\relax\relax sp\relax
}
\makeatother

\newcommand{\detv}[1]{\text{$\mathit{\lowercase{#1}}$}}
\newcommand{\detvmaj}[1]{\text{$\mathit{\uppercase{#1}}$}}

\newcommand{\knhds}[2]{\text{$\detv{\nu}_{#1}^{#2}$}}
\newcommand{\knlds}[2]{\text{$\detv{n}_{#1}^{#2}$}}
\newcommand{\qnx}[1]{\text{$\detvmaj{Q}_{\text{NX}}\left(#1\right)$}}
\newcommand{\rnx}[1]{\text{$\detvmaj{R}_{\text{NX}}\left(#1\right)$}}
\newcommand{\auc}{\text{AUC}}
\newcommand{\card}[1]{\text{$\left|#1\right|$}}
\usepackage{amsfonts} % for mathbb
\newcommand{\E}{\text{$\mathbb{E}$}} 
\newcommand{\Ep}[1]{\text{$\E\left[#1\right]$}} 

\begin{document}

\begin{frontmatter}

\title{SQuadMDS: a lean Stochastic Quartet MDS improving global structure preservation in neighbor embedding like $t$-SNE and UMAP}
% \title{Stochastic Quartet MDS (SQuadMDS): a lean MDS for better global structure preservation in neighbor embedding}

% that combine seamlessly

%% Group authors per affiliation:
\author{Pierre Lambert$^1$, Cyril de Bodt$^{1,2}$, Michel Verleysen$^1$, John A.~Lee$^{1,3}$}
\fntext[1]{UCLouvain - ICTEAM/ELEN. Place du Levant 3, bte L5.03.02, 1348 Louvain-la-Neuve, Belgium.}
\fntext[2]{MIT Media Lab - Human Dynamics group. 20 Ames Street, Wiesner Building (E15), Cambridge, MA 02139, USA.}
\fntext[3]{UCLouvain - IREC/MIRO. Avenue Hippocrate 55, B1.54.07, 1200 Brussels, Belgium.}

%% or include affiliations in footnotes:

%\author[mysecondaryaddress]{Global Customer Service\corref{mycorrespondingauthor}}
%\cortext[mycorrespondingauthor]{Corresponding author}
%\ead{support@elsevier.com}

%\address[1]{UCLouvain.be - ICTEAM/ELEN\\ 
%Place du Levant 3 L5.03.02, 1348 Louvain-la-Neuve - Belgium}
%\address[2]{UCLouvain.be - IREC/MIRO\\ 
%Avenue Hippocrate 55 B1.54.07, 1200 Brussels - Belgium }

\begin{abstract}
Multidimensional scaling is a statistical process that aims to embed high dimensional data into a lower-dimensional space; this process is often used for the purpose of data visualisation. Common multidimensional scaling algorithms tend to have high computational complexities, making them inapplicable on large data sets. This work introduces a stochastic, force directed approach to multidimensional scaling with a time and space complexity of $\mathcal{O}(N)$, with $N$ data points. The method can be combined with force directed layouts of the family of neighbour embedding such as $t$-SNE, to produce embeddings that preserve both the global and the local structures of the data.

Experiments assess the quality of the embeddings produced by the standalone version and its hybrid extension both quantitatively and qualitatively, showing competitive results outperforming state-of-the-art approaches. Codes are available at \url{https://github.com/PierreLambert3/SQuaD-MDS-and-FItSNE-hybrid}.

\end{abstract}

\begin{keyword}
Dimensionality reduction\sep $t$-SNE\sep UMAP\sep Neighbor embedding\sep MDS\sep Global structure\sep Stochastic algorithms\sep Data visualization\sep Distance preservation.
%\texttt{elsarticle.cls}\sep \LaTeX\sep Elsevier \sep template
%\MSC[2010] 00-01\sep  99-00
\end{keyword}

\end{frontmatter}

%\linenumbers

Multidimensional scaling (MDS) has been for long a popular method of embedding high-dimensional data nonlinearly in lower-dimensional spaces, in order to better visualize the proximities in data, for instance. 
This paper revisits the paradigm of MDS and distance preservation in general, by proposing a fast algorithm to compute gradients for a stochastic force-directed optimisation of the metric multidimensional scaling (mMDS) problem \cite{livreMDS}. An empirical evaluation of the method shows that it yields results that are often competitive with those obtained by the commonly-used SMACOF algorithm\cite{smacof}, while having linear time and memory complexities with respect to the number of observations instead of quadratic complexities. This work also shows that the stochastic mMDS gradients can be combined easily with gradients from neighbour embedding algorithms in order to produce solutions that both preserve the global and the local structures of the data. Public and free implementations are accessible at \url{https://github.com/PierreLambert3/SQuaD-MDS-and-FItSNE-hybrid}.

Section~\ref{sec:context} places mMDS in the current state of dimensionality reduction (DR) for data visualisation. Section~\ref{sec:prop_meth} describes the proposed method and its hybrid extension, which combines neighbour embeddings to mMDS. Section~\ref{sec:res} brings an empirical evaluation of both the standalone method and its hybrid extension.

\section{Context}
\label{sec:context}

The current state of DR for visualisation as well as the place of mMDS in this context is explained in Subsection~\ref{subsec:dr_vis}. 
The inherently high algorithmic complexity of mMDS is shown in Subsection~\ref{subsec:mMDS_chal}, as well as an overview of the currently available options to cope with this high complexity.

\subsection{Dimensionality reduction for visualisation}
\label{subsec:dr_vis}

Having evolved in a 3-dimensional environment, the human brain excels at representing concepts in a 2- or 3-dimensional space but struggles at understanding data that lies in a higher-dimensional (HD) space. Industry and academia generate an ever-increasing amount of HD data, which raises the question: how can we inspect and understand large volumes of data?

Dimensionality reduction is one way to address this problem; it is the process of embedding HD data into a lower-dimensional (LD) space such that some properties are preserved \cite{livreNLDR}. Having a visual representation of HD data can help us understand the dynamics of certain industrial processes \cite{chemIndustrial}, assist in monitoring \cite{monitoring}, or in the study of the relationships between certain products and consumer habits for marketing purposes \cite{cheeseDimensionalScaling}. DR for visualisation is also a particularly useful tool for researchers as the visual representations of HD data can help them develop intuition on their data and find relationships that might otherwise remain hidden \cite{retineTsne, tSNEchemistry}. When used for visualisation, the LD space is either 2- or 3-dimensional; for the sake of clarity, this paper will assume a LD space composed of 2 variables, although all developments can be carried out with larger numbers of LD features. 

Loss of information is often inevitable when reducing the dimensionality of the data. For this reason, many DR methods exist: each one focuses on the preservation of a certain property of the data. Some methods aim to preserve the local structures of the data by defining nonlinear pairwise similarities between the observations in HD and finding a LD embedding that preserves these similarities. Two popular examples are $t$-SNE \cite{vanDerMaaten2008} and UMAP \cite{umap}. While these algorithms excel at revealing the fine details of the data, the use of nonlinear pairwise similarities comes at a price: they tend to consider all the sufficiently dissimilar observations as equally dissimilar. This severely hinders the similarity-based methods capacity to capture the global properties of the data. It has been empirically shown that the global organisation of the embeddings produced by $t$-SNE and UMAP are reminiscent of the embedding initialisation, rather than a meaningful result of the optimisation process \cite{initIsKey}: two successive runs using different random initialisations will yield two global structures that are very likely to differ in the embeddings.

% ajout possible: si exploratory data analysis: on a pas ce expert knowledge. aussi, avec MDS on interprete mieux l embedding car on est dans un espace "plus euclidien" car le mapping de distances LD<->HD est plus linéaire.
When the global properties of the data are not retained in the embedding, the interpretation of the LD visualisation relies more heavily on the expert knowledge of the user. This can be dangerous as it encourages the users to see in the data what they think the data should show, with the risk of a confirmation bias. For this reason, when using DR to explore data, it can be desirable to use embeddings produced by algorithms that preserve large scale structures in conjunction to embeddings displaying information on the finer structures \cite{artOfTsne}. 

\begin{figure}[h]
    \centering
    \includegraphics[scale=0.4]{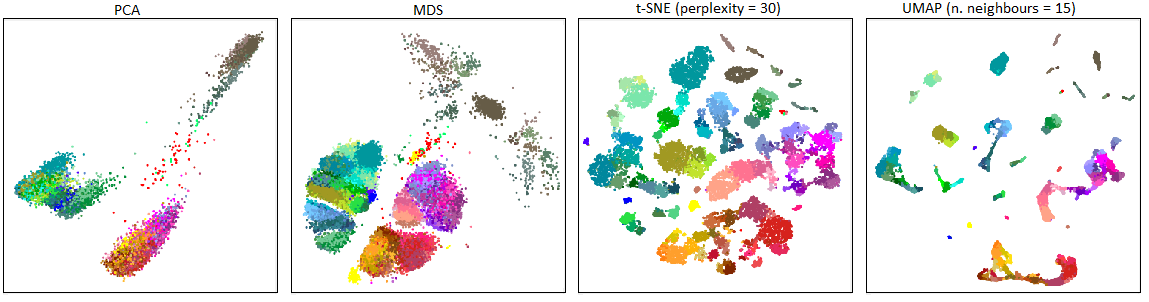}
    \caption{Embeddings of the mouse cortex data set.}
    \label{fig:4embeddings}
\end{figure}

Figure~\ref{fig:4embeddings} illustrates why relying on embeddings that preserve small-scale structures alone is problematic. The data set used in the figure comes from \cite{tasic}; each observation corresponds to a cell in a mouse brain, and the data was preprocessed as in \cite{artOfTsne}. The cells were coloured by the authors such that brown colours correspond to cells which aren't neurons, warm colours to inhibitory neurons, and cold colours to excitatory neurons.

The leftmost two embeddings were produced with PCA \cite{PCA} and mMDS \cite{livreMDS}; these tend to preserve the global structures in the data. The rightmost two embeddings were produced with $t$-SNE and UMAP, focusing on small-scale structures. The global structures present in the PCA and mMDS embeddings are lost in the rightmost two embeddings: if the user has no prior knowledge about the data, its hierarchical organisation would have remained hidden. This figure also shows that mMDS embeddings are more detailed and thus preferable to PCA embeddings for visualisation purposes. 

\subsection{Metric mutidimensional scaling and its challenges}
\label{subsec:mMDS_chal}

Metric multidimensional scaling is a statistical process that takes as input a distance matrix or HD data points and aims to embed these points into a LD space such that the Euclidean distances in the embedding are as close as possible to the distances in the HD space \cite{livreMDS}. Notice that the distances in HD are not necessarily Euclidean. As illustrated in Fig.~\ref{fig:4embeddings}, mMDS tends to produce embeddings that preserve the global structures of the data more than the local structures. 

The cost function for mMDS is often called "stress". In this work, the stress function is the mean squared error between the pairwise distances of the $N$ observations in HD and their LD counterparts. If $\delta$ denotes the distance matrix in HD, $d$ denotes the Euclidean distance matrix in LD, and  $\Delta_{ij} = \delta_{ij} - d_
{ij}$, then the objective of mMDS is to minimise

\[
    \mathlarger{\mathlarger{\mathcal{L}^{mMDS}}} = \mathlarger{\mathlarger{\frac{1}{{N \choose 2}}} }\; \mathlarger{\mathlarger{\sum\limits_{i < j}}}\;\;  \Delta_{ij}^2 \enspace .
\]

One of the most common algorithms to perform mMDS is the \mbox{SMACOF} algorithm (Scaling by MAjorizing a COmplicated Function) \cite{smacof}: it involves an iterative optimisation with iterations of $\mathcal{O}(N^2)$ time complexity. This relatively high computational requirement stems from the fact that the \mbox{SMACOF} algorithm minimises the exact pairwise formulation of the stress function, which has ${N \choose{2}}$ terms in the summation. The volume of data that is being accumulated nowadays far outpaces Moore's law, making such a high computational complexity problematic. 

% trouver plus belle phrase pour "sampling observations from what is already a sample of observations"
The high computational complexity of mMDS often forces the visualisation community to perform mMDS on a random subsample of the dataset or on the prototypes resulting from a clustering or vector quantization of the data in HD space \cite{artOfTsne}. This can be dangerous as it brings a risk of over- or under-representing some parts of the data, which would lead to embeddings that poorly reflect the true data distribution. 

A large number of data sets have a number of observations $N$ that is much larger than the number of dimensions $M$, which implies that the original data matrix has fewer entries than the HD distance matrix ($N\times M$ entries compared to $\frac{N(N-1)}{2}$ effective entries). This can lead to the intuition that the higher the ratio $\frac{N}{M}$, the more the HD distance matrix contains redundant information. As a consequence, the full distance matrix is not necessary to find a good mMDS solution. Faster mMDS algorithms have been designed building directly or indirectly on this intuition; the next paragraph presents a brief overview of some of these algorithms.

Morrison et al.~\cite{Morrison} propose an iterative multi-staged force-directed layout with a temporal complexity of $\mathcal{O}(N\sqrt{N})$; this algorithm places the bulk of the data points around an initial embedding of a small sample of the data set. Silva et al.~follow a comparable approach in landmark-MDS \cite{landmark}. Yang et al.~propose a divide-and-conquer approach to mMDS \cite{fastMDS} by dividing the HD distance matrix along the diagonal into smaller matrices. The small distance matrices are then independently used to embed the corresponding points; afterwards, the multiple embeddings are stitched together for a total time complexity of $\mathcal{O}(N\log N)$. Another interesting approach is by Williams and Munzner \cite{steerable}, where the high computational complexity of mMDS is mitigated by asking the input of a human supervisor to focus the computations on specific parts of the embedding.

While these algorithms significantly reduce the computational resources necessary to compute a usable mMDS embedding, they come with their own set of limitations. The multiple steps required by these algorithms often entail some additional hyperparameters, which need to be adjusted by the user, depending on the data set and the purpose of the visualisation, a potentially long process. Some multi-staged algorithms such as the divide-and-conquer approach also become intractable black-boxes to the user, who needs to wait for the completion of the algorithms to get a visual feedback. This can be detrimental to the process of data visualisation which often involves interactivity between the user and the data. Finally, even algorithmic complexities of $\mathcal{O}(N\log N)$ or $\mathcal{O}(N\sqrt{N})$ can become demanding when the data sets grow very large.

This work attempts to address these shortcomings by proposing an algorithm that perform mMDS in $\mathcal{O}(N)$ time and space complexity, using a simple stochastic gradient descent (SGD) optimisation. Assuming a target dimensionality of 2, the base method involves no other hyperparameters than the maximum number of iterations for the SGD optimisation. 
The simplicity of the proposed method also enables it to be mixed easily with neighbour-embedding methods, as a hybridization between the distance-scaling paradigm and the similarity-preservation paradigm. The purpose of the hybrid approach is to counter the difficulties of the similarity-based methods such as $t$-SNE and UMAP to preserve large-scale structures, by allowing the distance-scaling gradients to guide the optimisation towards a representative global organisation.
%which produces embeddings that both preserve the local and the global structures of the data. 

\section{Proposed method}
\label{sec:prop_meth}

%This section introduces the proposed method to optimise mMDS stress function by SGD.
The proposed method performs mMDS by using SGD to optimise a stress function. This section details the standalone mMDS method as well as the hybrid extension, which uses gradients from both the distance-scaling and the similarity-preservation paradigms. 

Subsection~\ref{subsec:comp_grad} introduces the standalone version of the method, describing the overall SGD strategy and how the gradients are computed at each iteration. Subsection~\ref{subsec:rel_dist} explains the motivations for using relative distances instead of absolute distances. Subsection~\ref{subsec:hybrid_ext} proposes a scheme to blend the mMDS gradients with those of $t$-SNE, a similarity-based DR algorithm which is also optimised through gradient descent.

\subsection{Computing the gradients}
\label{subsec:comp_grad}

To keep the computational complexity of each SGD iteration low, this works takes a divide-and-conquer approach to compute the gradients. At each iteration, the data set is randomly partitioned into groups of 4 points called quartets \cite{lambert2021stochastic}. The previously stated mMDS stress function is redefined to put independent quartets of points to the forefront, in order to compute the gradients for each point by looking only at the three other points in the quartet at each SGD iteration.
%The gradients are computed from a loss function defined independently on each quartet of points.
The proposed method has been coined "SQuaD-MDS", standing for "Stochastic Quartet Descent MDS".

The motivation behind the use of quartets of points comes from triangulation: in a 2-dimensional plane, a point can be accurately placed given its distances to three other points and their locations. This work assumes a target dimensionality of 2 but adapting it to a target dimensionality of 3 would only require using groups of 5 points instead of 4. 

Let us start by rewriting the previously defined stress function of mMDS in a way that highlights quartets of points:
\begin{equation}\label{eqn:bigloss}
    \mathlarger{\mathlarger{\mathcal{L}^{mMDS}}} \;\;=  \;\;
    \mathlarger{\mathlarger{\frac{1}{{N \choose 2}}} }\; \mathlarger{\mathlarger{\sum\limits_{i < j}}}\;\;  \Delta_{ij}^2\;\; =\;\;
    \mathlarger{\mathlarger{\frac{1}{{N \choose 4} }}}\; \mathlarger{\mathlarger{\sum\limits_{i < j < k < l}}\;} \mathlarger{\mathlarger{\frac{1}{{4 \choose 2}} \;\mathcal{L}_{ijkl}^{Q}}} \tag{1} \enspace ,
\end{equation}
where $\mathcal{L}_{ijkl}^{Q}$ is the stress term for a single quartet composed of the points indexed by $i$, $j$, $k$, and $l$; that is, 
%Let us start by defining a new stress function for mMDS which is based on quartets of points: 
%\begin{equation}\label{eqn:bigloss}
%    \mathlarger{\mathlarger{\mathcal{L}^{mMDS}}} \;\;=  \;\;
%    \mathlarger{\mathlarger{\frac{1}{{N \choose 4} } \; \sum\limits_{\mathcal{I}_q\in\mathcal{P}_4^N}\;\frac{1}{{4 \choose 2}}}}\; \mathlarger{\mathlarger{\sum\limits_{\left\{i,j,k,l\right\}\in\mathcal{I}_q}}\;} \mathlarger{\mathlarger{\mathcal{L}_{ijkl}^{Q}}} \tag{1} \enspace ,
%\end{equation}
%where $\mathcal{P}_4^N$ is the set of all partitions of $\left\{1,2,\hdots,N\right\}$ into subsets of 4 elements, $\mathcal{I}_q$ is one of these partitions which identifies one assignment of the data points into quartets, and $\mathcal{L}_{ijkl}^{Q}$ is the genuine stress term for a single quartet composed of the points indexed by $i$, $j$, $k$, and $l$, that is, 
\[
    \mathlarger{\mathcal{L}_{ijkl}^{Q} =\;\; \Delta_{ij}^2\;+\;\Delta_{ik}^2\;+\;\Delta_{il}^2\;+\;\Delta_{jk}^2\;+\;\Delta_{jl}^2\;+\;\Delta_{kl}^2\;} \enspace .
\]
Each possible quartet of data points appears exactly once in the right hand side of~\eqref{eqn:bigloss}. 
Such a formulation is particularly well suited to the deployment of stochastic gradient descent, by considering the quartets separately during the optimization process. 

Instead of deriving the gradients directly from $\mathcal{L}_{ijkl}^{Q}$, SQuaD-MDS uses an alternate version of the quartet stress that uses relative distances, $\mathcal{L}_{ijkl}^{Q_{rel}}$. The relative distances are defined as being the absolute distance divided by the sum of the six distances that appear in the quartet. The relative distance between the $i^\text{th}$ and $j^\text{th}$ points belonging to the quartet composed of the points indexed by $i$, $j$, $k$, and $l$ are noted $\delta^{rel}_{\bm{ij}kl}$ and $d^{rel}_{\bm{ij}kl}$ in HD and LD, respectively 
\small
\[
\delta^{rel}_{\bm{ij}kl} \; = \; \frac{\delta_{ij}}{{\delta_{ij} + \delta_{ik} + \delta_{il} + \delta_{jk} + \delta_{jl} + \delta_{kl}}} \;\;\text{;}\;\;\;\;d^{rel}_{\bm{ij}kl} \; = \; \frac{d_{ij}}{{d_{ij} + d_{ik} + d_{il} + d_{jk} + d_{jl} + d_{kl}}} \enspace .
\nonumber
\]
\normalsize

The quartet stress term using relative distances simply replaces the absolute distances in $\mathcal{L}_{ijkl}^{Q}$ with relative distances:
\[
    \mathlarger{\mathcal{L}_{ijkl}^{Q_{rel}} =\;\; {\Delta_{\bm{ij}kl}^{{rel}}}^2\;+\;{\Delta_{\bm{ik}jl}^{{rel}}}^2\;+\;{\Delta_{\bm{il}jk}^{{rel}}}^2\;+\;{\Delta_{\bm{jk}il}^{{rel}}}^2\;+\;{\Delta_{\bm{jl}ik}^{{rel}}}^2\;+\;{\Delta_{\bm{kl}ij}^{{rel}}}^2\;} \enspace ,
\]
where
\[
\mathlarger{\Delta_{\bm{ij}kl}^{{rel}} \; = \; \delta^{rel}_{\bm{ij}kl} - d^{rel}_{\bm{ij}kl} } \enspace .
\]

%The stress term $\mathcal{L}_{ijkl}^{Q_{rel}}$ is derivable with respect to each LD coordinate in the quartet. By naming $\mathbf{I}$ the $N$-by-$N$ identity matrix, $x_q$ the value taken by one of the LD axis for the $q^{\text{th}}$ point with $q \in \{i, j, k, l\}$, and $S = d_{ij} + d_{ik} + d_{il} + d_{jk} + d_{jl} + d_{kl}$, the gradient for one element of the sum in $\mathcal{L}_{ijkl}^{Q_{rel}}$ is
%
%\small
%\begin{eqnarray}
%\dfrac{\partial({\Delta_{\bm{ij}kl}^{{rel}}}^2)}{\partial x_q}
%=
%\frac{2(d^{rel}_{\bm{ij}kl} - \delta^{rel}_{\bm{ij}kl})}{S}
%\bigg(
%\xoverbrace{\frac{\mathbf{I}_{qi}(x_q-x_j) + \mathbf{I}_{qj}(x_q-x_i)}{d_{ij}}}^{\text{\circled{A}}}
%\;\;\xoverbrace{- \; d^{rel}_{\bm{ij}kl} \mathlarger{\sum\limits_{ \substack{b \in \{i,j,k,l\}  \\ b \neq q}} }\frac{x_q - x_b}{d_{qb}}
% }^{\text{\circled{B}}}
%\bigg) \enspace .
%\nonumber
%\end{eqnarray}
%\normalsize
%\newline

The stress term $\mathcal{L}_{ijkl}^{Q_{rel}}$ is derivable with respect to each LD coordinate in the quartet. By naming $\mathbf{I}$ the $N$-by-$N$ identity matrix, $\mathbf{x}_q$ the LD coordinates of the $q^{\text{th}}$ point with $q \in \{i, j, k, l\}$, and $S = d_{ij} + d_{ik} + d_{il} + d_{jk} + d_{jl} + d_{kl}$, the gradient for one element of the sum in $\mathcal{L}_{ijkl}^{Q_{rel}}$ is
\small
\begin{eqnarray}
\dfrac{\partial({\Delta_{\bm{ij}kl}^{{rel}}}^2)}{\partial \mathbf{x}_q}
=
\frac{2(d^{rel}_{\bm{ij}kl} - \delta^{rel}_{\bm{ij}kl})}{S}
\bigg(
\xoverbrace{\frac{\mathbf{I}_{qi}(\mathbf{x}_q-\mathbf{x}_j) + \mathbf{I}_{qj}(\mathbf{x}_q-\mathbf{x}_i)}{d_{ij}}}^{\text{\circled{A}}}
\;\;\xoverbrace{- \; d^{rel}_{\bm{ij}kl} \mathlarger{\sum\limits_{ \substack{b \in \{i,j,k,l\}  \\ b \neq q}} }\frac{\mathbf{x}_q - \mathbf{x}_b}{d_{qb}}
 }^{\text{\circled{B}}}
\bigg) \enspace .
\nonumber
\end{eqnarray}
\normalsize

\circled{A} is zero if $q\not\in\{i, j\}$, this part of the gradient directly moves the point towards or further from the $i^{\text{th}}$ or $j^{\text{th}}$ point, similarly to the forces that would be exerted using gradients derived from a stress function using absolute distances. 

\circled{B} comes from the relative distances. This part of the equation is a more global term that influences the quartet stress by changing the denominator in the definition of the relative distances. For instance, if $q\not\in\{i, j\}$ and $d^{rel}_{\bm{ij}kl}$ is too small compared to $\delta_{\bm{ij}kl}^{{rel}}$, then the $q^{\text{th}}$ point will be attracted to the $i^{\text{th}}$ and $j^{\text{th}}$ points in order to increase $d^{rel}_{\bm{ij}kl}$ by decreasing its denominator.

The gradients for each point are computed in constant time relative to the number of observations $N$ and each point receives an update at each iteration; the time complexity of each SGD iteration is therefore $\mathcal{O}(N)$. The full HD and LD distance matrices are not necessary as the distances can be computed on the fly when the quartets are designated; the spatial complexity of the proposed method is therefore also low with respect to $N$. % It can be interesting to notice that for large datasets, the number of considered distances throughout the full optimisation is much smaller that the size of the upper triangle of the distance matrix ($ \text{number\_of\_iterations} \times6\times \lfloor\frac{N}{4}\rfloor$ versus $\frac{N(N-1)}{2}$).\\

So far, this subsection has explained how the proposed method randomly partitions the data into quartets of points at each SGD iteration. Once the data is organised into quartets, the gradients are derived from the quartet-specific cost function $\mathcal{L}_{ijkl}^{Q_{rel}}$. These gradients, however, are not applied directly to the LD coordinates; the remaining of this subsection introduces the specificities of the SGD strategy taken by SQuaD-MDS. 

The partitioning into quartets at each iteration is a sampling process where $\lfloor\frac{N}{4}\rfloor$ non-overlapping quartets are sampled from all the ${N \choose 4}$ possible combinations of quartets. Because of the stochastic nature of this process, the learning rate is decayed following ${(a\cdot t + b)}^{-1}$ with $a$ and $b$ as constants and $t$ the iteration number, in accordance to the Robbins-Monro principles for stochastic optimisation. 
%exponentially during the optimisation. The optimisation process is quite robust to changes in the decay rate as long as the learning rate diminishes by about $3$ or more magnitudes by the last SGD iteration. 

Additionally, the optimisation updates each coordinate by keeping track of their momentum. The intuition motivating this approach is that the momenta capture a larger part of the sum of quartet costs in~\eqref{eqn:bigloss}, bringing updates that better follow the global energy potentials than if using standalone gradients. This makes the optimisation process more robust with respect to randomness. 

In practice, this work uses an improved version of the momentum introduced by Yurii Nesterov in \cite{NesterovUSSR}. 
The Nesterov method first computes temporary coordinates by updating the current coordinates with their momentum; the gradients are computed using these temporary coordinates. The gradients are then used to update the momenta; finally, the real coordinates are changed by following the updated momenta. This process is described below, with $\theta_t$ and $v_t$ a parameter at iteration $t$ and its momentum, $\nabla_\theta J(\theta)$ the gradient of the cost function with respect to the parameter $\theta$, and $\eta$ and $\gamma$ constants balancing the weight of the momentum relatively to the gradient:
\small
\begin{align*}
v_t&=\gamma\cdot v_{t-1}-\eta \nabla_\theta J(\theta_{t-1}+\gamma\cdot v_{t-1}),\\
\theta_t&=\theta_{t-1}+v_t.
\end{align*}
\normalsize
The Nesterov method for gradients brings faster convergence in a non-stochastic context. Despite the fact that the proposed optimisation is a stochastic process, the usage of Nesterov momenta improves the quality of the results while only requiring a negligible supplementary cost in terms of computations. 
%The method computes the gradients using parameters that are already updated by their momentum, this brings faster convergence in a non-stochastic context. 

Some gradient descent methods add an additional sophistication to the optimisation by using the Hessian matrix or an approximation of the matrix to determine norm of the gradients \cite{LBFGS}. This can yield efficient updates in a non-stochastic context, but at the cost of additional computations. This work relies on a large number of fast iterations cemented together by a strong momentum; computing an approximation of the Hessian matrix at each iteration would significantly slow down the algorithm and risk propagating the noise from the stochastic environment to the Hessian matrix. For this reason, the SGD process proposed in this work does not adopt principles used in second order methods.
%Gradient descent and its stochastic version are very active research topics and many optimisation strategies exist. This work favours the momentum approach for its simplicity and to reduce the time to compute the gradients. Some more sophisticated approaches might even be detrimental to the global quality of the results because of the strong stochastic nature of the optimisation. For instance, looking at the Hessian matrix to determine the step size could risk giving too much power to the often unavoidable minority of poorly placed points, destabilizing the embedding.

\subsection{Motivation for the use of relative distances}
\label{subsec:rel_dist}

The main reason behind the usage of relative distances is to remove the implicit scale constraint that appears when preserving the absolute distances. Constraining the scale of the embedding is deemed unnecessary as the LD space is composed of abstract variables that are not directly observable: the information in the embedding is carried by the relative positions of the points, not by their absolute distances. Not having a scale constraint is also useful to evade conflicts concerning the scale when mixing the SQuaD-MDS gradients with those of other DR methods as described in the following subsection.

\subsection{Hybrid extension}
\label{subsec:hybrid_ext}

Since this work optimises a distance preserving cost function using SGD, the gradients can be blended with those of other DR methods that also rely on gradient descent. The objective of such a hybridisation is to produce embeddings that minimise multiple cost functions for better overall quality. This work suggests mixing SQuaD-MDS's gradients with those of $t$-SNE \cite{vanDerMaaten2008}. The motivation behind the choice of $t$-SNE for hybridisation is that $t$-SNE and mMDS have complementary objectives: mMDS focuses on global structures and $t$-SNE on local structures.

The exact version of $t$-SNE has a high time complexity of $\mathcal{O}(N^2)$; blending it with fast mMDS gradients would defeat the purpose of this work entirely. For this reason, we choose FI$t$-SNE: an accelerated approximation of $t$-SNE introduced by Linderman, Kluger, et al.~in \cite{fitsne}. FI$t$-SNE uses gradient descent with iterations of $\mathcal{O}(N)$ time complexity, the same complexity as SQuaD-MDS. The details of $t$-SNE and FI$t$-SNE are outside of the scope of this paper and are detailed in \cite{vanDerMaaten2008} and \cite{fitsne}. 

In order to combine both methods, this work keeps the same protocol as for the standalone version of SQuaD-MDS: the learning rates are decayed and Nesterov's momentum is used. 

Since the FI$t$-SNE and SQuaD-MDS gradients are computed using different cost functions, there is no guarantee that their norms are of comparable magnitude. This is problematic when aiming to mix the gradients together, as one gradient type could overwhelm the effect of the other type by having naturally bigger norms. In order to reduce the difference in magnitude between the gradients of both type, the standard deviation of the norm of the SQuaD-MDS gradients and FI$t$-SNE gradients are computed independently over the LD points. The gradients of each type are then divided by the corresponding standard deviations as a form of normalisation. Once the gradients normalised, learning rates individual to each method are used to combine the gradients. Changing the balance between the learning rates of FI$t$-SNE and SQuaD-MDS enables the user to prioritize one paradigm over the other, focusing on either local or global preservation. This work suggests initial learning rate values of $1$ for FI$t$-SNE and $0.5$ for SQuaD-MDS as default values, based on empirical experiments. 

$t$-SNE and FI$t$-SNE require a crucial scalar hyperparameter called perplexity. A simplistic but sufficient way to view the perplexity here is to consider it as the rough number of neighbours that will be considered when embedding the points: a small perplexity will produce an embedding that preserves well the close neighbours of each point and a large perplexity will consider larger sets of neighbours. A high perplexity is not always preferable as it comes at a price in terms of preservation of small-scale structures.
%the sets of $K$-nearest neighbours in HD and LD with \mbox{$K << $} perplexity have a smaller normalised intersection than when \mbox{$K \approx$ perplexity}. 

The purpose of FI$t$-SNE's gradients in the hybridisation with SQuaD-MDS is to preserve the small-scale structures. For this reason, this work uses multi-scale similarities \cite{RNX,ttsne_mstsne} limited to small perplexities. As a default, this work suggests perplexities equal to $4$ and $50$. These values were determined empirically; for very large data sets, the perplexity of $50$ may be slightly increased for better results. 

\section{Empirical quality assessment}
\label{sec:res}

Qualitative and quantitative  assessments of SQuaD-MDS and its hybrid extension are presented in this section. 

Subsection~\ref{subsec:dataRnx} introduces the data sets used for the evaluation and brings a short introduction to $R_\mathrm{NX}(K)$ curves \cite{RNX}, which are used to assess the quality of the neighbour preservation across all scales. 
In order to evaluate the performance of the standalone SQuaD-MDS algorithm, Subsection~\ref{subsec:resultsMDS} compares its results to mMDS embeddings as yielded by the SMACOF algorithm. Subsection~\ref{subsec:resultsHybrid} assesses the quality of embeddings obtained with the hybrid approach by comparing them to embeddings produced by FI$t$-SNE and UMAP.

\subsection{Data sets and quality curves}
\label{subsec:dataRnx}

Eight data sets are featured in this paper, with size $N$ and dimensionality $M$. MNIST digits: ($N$, $M$) = ($60000$, $784$); MNIST fashion: ($N$, $M$) = ($60000$, $784$); gesture phase segmentation (gestures) \cite{gestures}: ($N$, $M$) = ($9901$, $784$); airfoil self-noise (airfoil): ($N$, $M$) = ($1502$, $5$); Boston housing (housing): ($N$, $M$) = ($404$, $13$); Statlog landsat satellite: ($N$, $M$) = ($4434$, $36$); single-cell RNA-seq data from the adult mouse cortex (mouse cortex) \cite{tasic}: ($N$, $M$) = ($23822$, $50$); single-cell RNA-seq data of a zebrafish embryo coloured by time after fertilisation (zebrafish) \cite{zebrafish}: ($N$, $M$) = ($63530$, $50$). 

Both single-cell data sets are composed of their $50$ principal components after going through the preprocessing pipeline described in \cite{artOfTsne}. Most of the other data sets are available on the UCI machine learning data repository \cite{Dua_2019}. 

%The quality of the embeddings is assessed by using $R_\mathrm{NX}(K)$ curves \cite{RNX}. 
%The $R_\mathrm{NX}(K)$ curves rely on the measure of the mean intersection between the HD and LD $K$-ary neighbour sets of each point on every scale (from $K = 1$ to $K = N-1$). 
%Specific baseline removal and renormalization make that $R_\mathrm{NX}(K)$ has values close to 1 for a certain scale $K$ if the preservation of neighbourhoods is nearly perfect on this particular scale, while values near zero mean that the neighbourhoods are not preserved on this scale better than a random embeddings would have done on average. 
%The curves are shown using a logarithmic scale for scale $K$, as the large scales would predominate in the diagram otherwise. 
%The $R_\mathrm{NX}(K)$ curves \cite{RNX} are accompanied by their area under the curve (AUC); this scalar number is a way to assess the overall quality of an embedding: the higher the value, the better the general preservation of neighbourhoods. 
%The AUC is also based on a logarithmic scale for K. 

The HD neighborhood reproduction in the LD embedding is deemed as driving the DR quality in numerous studies \cite{venna2010information, chen2009local, cdb2018drnap, mokbel2013visualizing}. 
Some criteria have been developed to quantify this reproduction \cite{lee2009quality,lee2010scale}. 
One can first denote as $\knhds{i}{K}$ and $\knlds{i}{K}$ the $K$ nearest neighbors of the $i^{\text{th}}$ data point, respectively in the HD and LD spaces. 
These $K-$ary neighborhoods have an average normalized intersection developing as 
\begin{equation*}
\qnx{K} = \frac{1}{N} \sum\limits_{i=1}^N \frac{\card{\knhds{i}{K} \cap \knlds{i}{K}}}{K} \in \left[0,1\right]. 
\end{equation*}
As defining the LD coordinates randomly leads to $\Ep{\qnx{K}}=K/\left(N-1\right)$, 
\begin{equation*}
\rnx{K} = \frac{\left(N-1\right)\qnx{K}-K}{N - 1 - K}
\end{equation*}
is an indicator that allows to compare neighborhoods with varying sizes \cite{lee2013type}. 
Since the literature generally believes local neighborhoods as more important \cite{vanDerMaaten2008,cdb2019catsne,roweis2000nonlinear} and as larger scales outnumber smaller ones by far, $\rnx{K}$ can be plotted using a log-scale for $K$. The area under the obtained curve ($\auc$), 
\begin{equation*}
\auc = \left(\sum_{K=1}^{N-2}K^{-1}\right)^{-1}\cdot\sum_{K=1}^{N-2}\frac{\rnx{K}}{K} \in\left[-1,1\right],
\label{eq:def_aucnx}
\end{equation*}
quantifies DR quality by considering all scales, focusing on smaller ones preferentially \cite{RNX,fast_ms_NE}. 
Greater (resp. smaller) values indicate better (resp. lower) quality. 

\subsection{Assessment of the standalone SQuaD-MDS}
\label{subsec:resultsMDS}

\begin{center}
    \begin{figure}[H]
        \centering
        \includegraphics[scale=0.4]{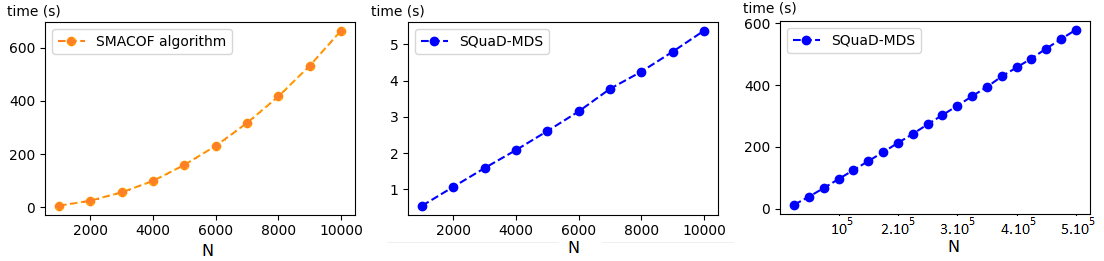}
        \caption{Time taken by the SMACOF algorithm and SQuaD-MDS to yield an embedding, as a function of the data set size $N$.}
        \label{fig:time}
    \end{figure}
\end{center}

An empirical evaluation of the time taken by $5000$ iterations of SQuaD-MDS across multiple data set sizes $N$ is shown in Fig.~\ref{fig:time}. The time taken by the SMACOF algorithm using the default set of hyperparameters proposed by the \textit{scikit-learn} library in Python \cite{scikit-learn} is shown on the leftmost plot. 

This figure shows that the proposed method took less than $6$ seconds to produce an embedding when $N=10^4$, whereas the SMACOF algorithm required more than $600$ seconds in the same testing environment. The linear time complexity of SQuaD-MDS with regards to $N$ is apparent on the rightmost plot.

\begin{figure}[H]
    \centering
    \includegraphics[scale=0.35]{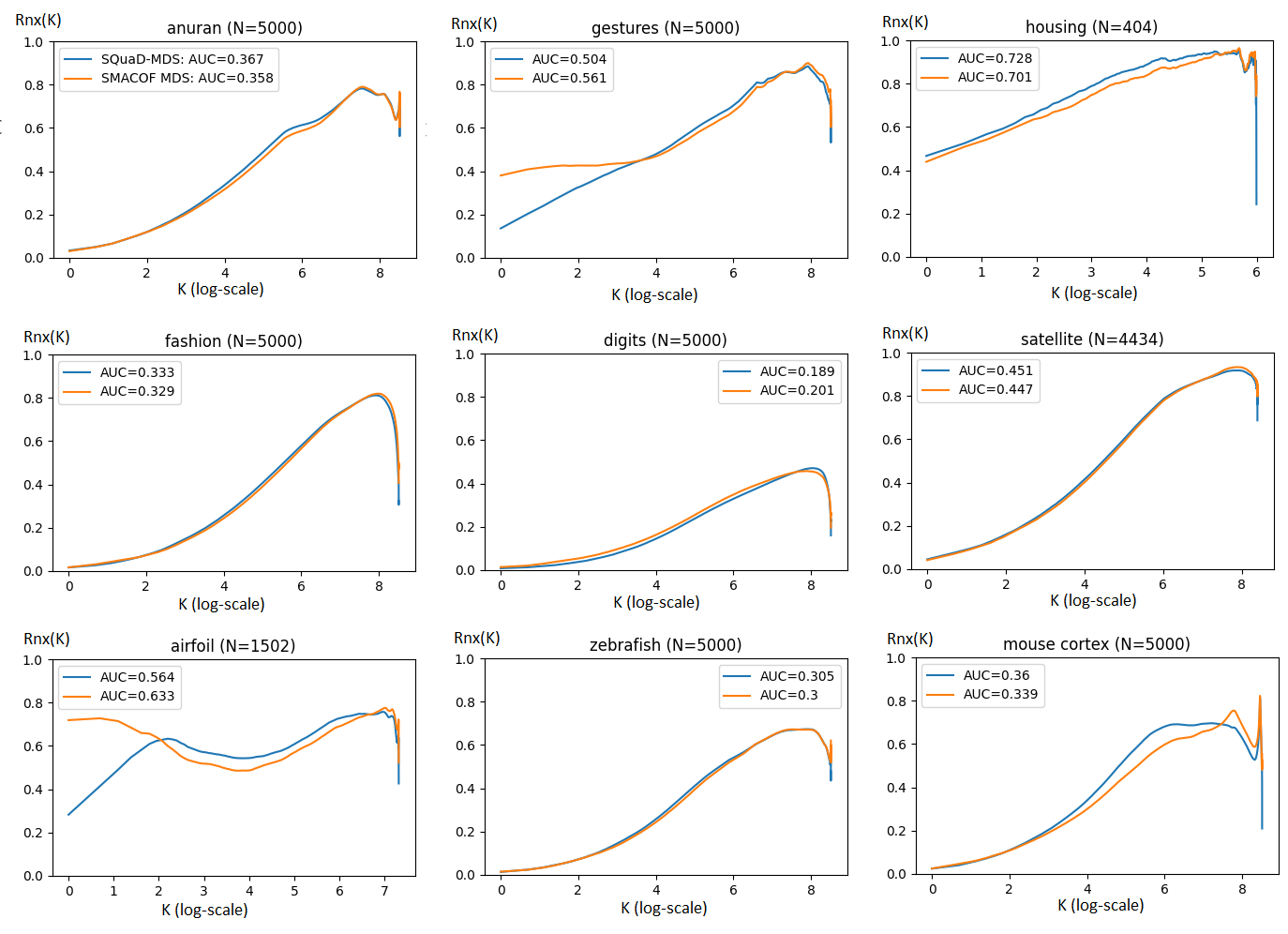}
    \caption{$R_\mathrm{NX}(K)$ curves of embeddings produced by the SMACOF algorithm and SQuaD-MDS.}
    \label{fig:MDSrnx}
\end{figure}

\begin{figure}[H]
    \centering
    \includegraphics[scale=0.4]{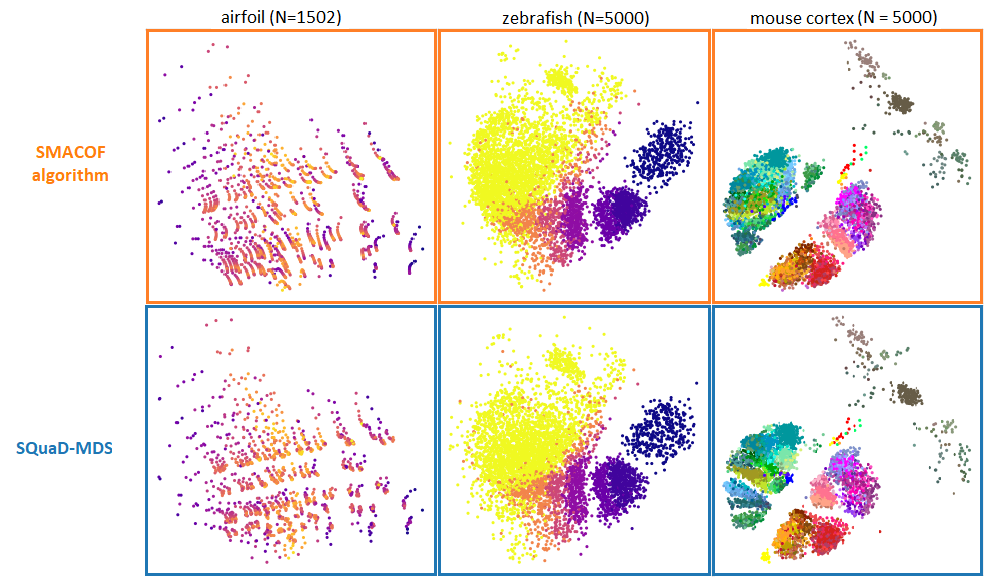}
    \caption{Embeddings produced by the SMACOF algorithm and SQuaD-MDS.}
    \label{fig:MDSembeddings}
\end{figure}

The quality of embeddings produced by the SMACOF algorithm and SQuaD-MDS can be compared in Figs.~\ref{fig:MDSrnx} and~\ref{fig:MDSembeddings}. The data sets larger than $5000$ points were sampled down to $5000$ in order to cope with the high computational complexity of the SMACOF algorithm. The number of iterations for the SQuaD-MDS embeddings and the hyperparameters of the SMACOF algorithm are the same as for the time evaluation in Fig.~\ref{fig:time}. 

The $R_\mathrm{NX}(K)$ curves of Fig.~\ref{fig:MDSrnx} show that both methods tend to favor the preservation of global structures (high $R_\mathrm{NX}(K)$ values on large $K$). The AUC are often similar for both methods, although in the gestures and airfoil data sets the SMACOF algorithm performs better at preserving small-scale structures than SQuaD-MDS. 

This can be explained by the fact that SQuaD-MDS is stochastic and largely subsamples the population of all possible quartets, neglecting many of the small-area quartets, which leaves some noise in the shortest distances and hence also in the final coordinates of the data points. This phenomenon can be qualitatively observed in Figure~\ref{fig:MDSembeddings} in the airfoil embeddings: the fine $1$-dimensional manifolds observed in the SMACOF embedding are not as delicate in the SQuaD-MDS embedding. The shortcomings of SQuaD-MDS in terms of small-scale structures with some data sets motivate the hybrid approach with $t$-SNE. The weakness of SQuaD-MDS concerning the preservation of fine structures on certain data sets is limited in practice, as the purpose of mMDS is generally to explore the global organisation of the data more than the local structures.

\subsection{Assessment of the hybrid extension}
\label{subsec:resultsHybrid}

The hybrid method combines the distance scaling paradigm to the neighbour embedding paradigm by using gradients from SQuaD-MDS and FI$t$-SNE. In order to assess the quality of the embeddings produced by the hybrid method, they are compared to those produced by FI$t$-SNE \cite{fitsne} and UMAP \cite{umap}; both are popular DR methods focusing on the small-scale structures in the data. 

For the sake of comparison, all the methods use the same PCA initialization of the embedding. 

The hybrid method was applied using the proposed default hyperparameters, notably perplexity values of $4$ and $50$ to compute the multi-scale similarities. The number of iterations of the optimisation was set to $750$, which is the default value in the implementation of FI$t$-SNE made by the authors. 

For the sake of comparison, the FI$t$-SNE embeddings were also produced by combining perplexities of $4$ and $50$; all other hyperparameters were kept to their default values. 

The UMAP method builds on different principles than $t$-SNE and its accelerated versions; for this reason, the choice of hyperparameters to conduct a comparison is not obvious. Similarly to the perplexity, UMAP takes a hyperparameter called the number of neighbours; it is responsible of the balance between the preservation of local and global structures in the data. Despite having the same purpose, the number of neighbours used by UMAP does not translate exactly to the perplexity used by $t$-SNE in terms of effective balance between small-scale and large-scale structures. To produce UMAP embeddings of quality while achieving a similar trade-off between the preservation of small-scale and large-scale structures as in the other methods, the following experiments apply UMAP ten times per data set with a number of neighbours taking values in ${5, 10, 15, ..., 55}$. From the ten embeddings, the one with the best AUC of the $R_\mathrm{NX}(K)$ curve is kept and considered in the results. 

\begin{center}
    \begin{figure}[H]
        \centering
        \includegraphics[scale=0.35]{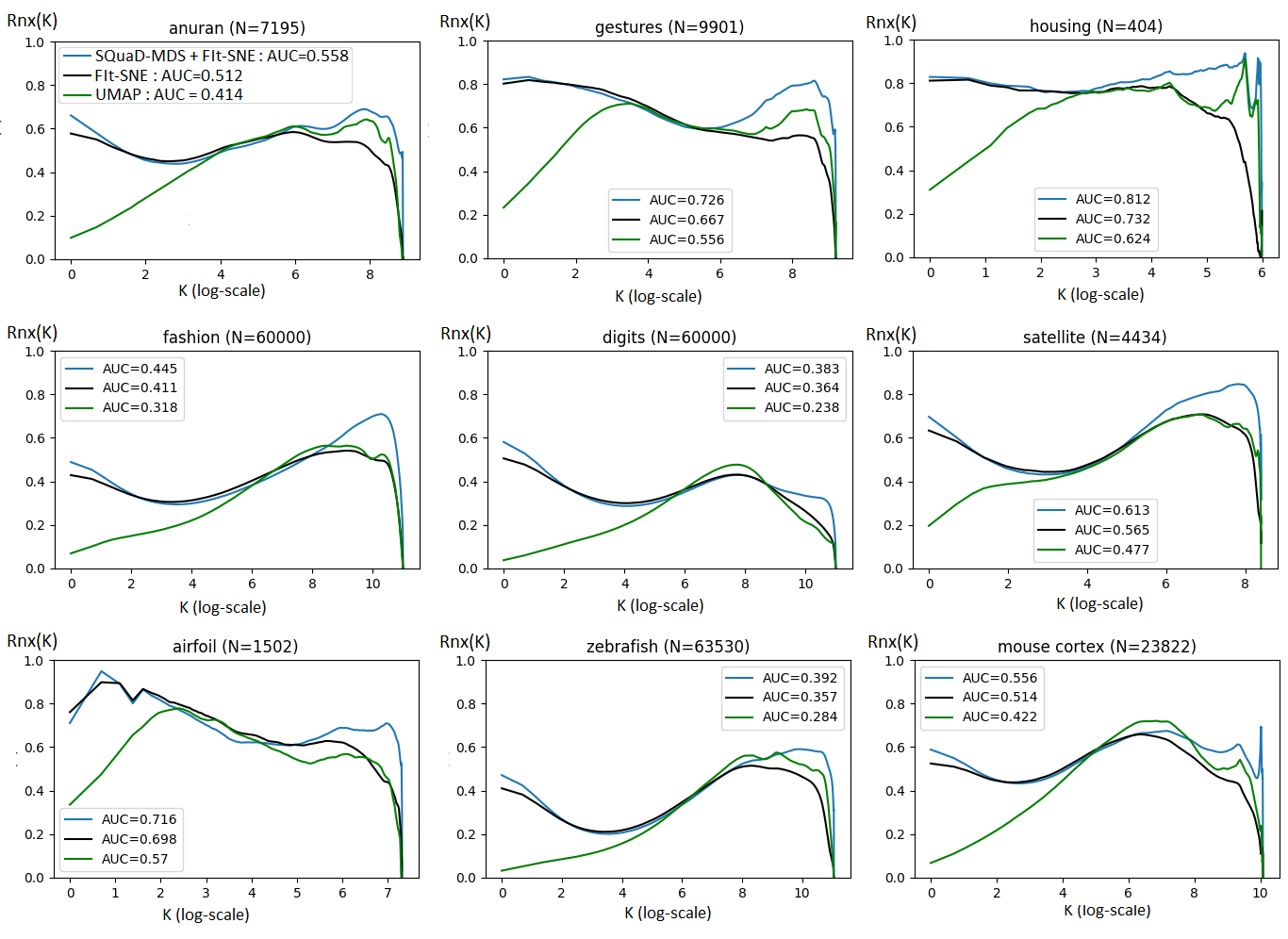}
        \caption{$R_\mathrm{NX}(K)$ curves of embeddings produced by the proposed hybrid method, FI$t$-SNE, and UMAP.}
        \label{fig:rnx_hybrid}
    \end{figure}
\end{center}

The $R_\mathrm{NX}(K)$ curves presented in Fig.~\ref{fig:rnx_hybrid} show a comparison of the quality of the embeddings produced by the three methods across multiple scales. On the tested data sets, the hybrid method has similar $R_\mathrm{NX}(K)$ values to those of FI$t$-SNE on the small and medium scales, but its scores on large $K$ values are higher. This shows that the combination of distance preserving gradients to those of FI$t$-SNE can enhance the preservation of the large-scale structures of the data, without impeding on the preservation of the small-scale structures.

\begin{center}
    \begin{figure}[H]
        \centering
        \includegraphics[scale=0.4]{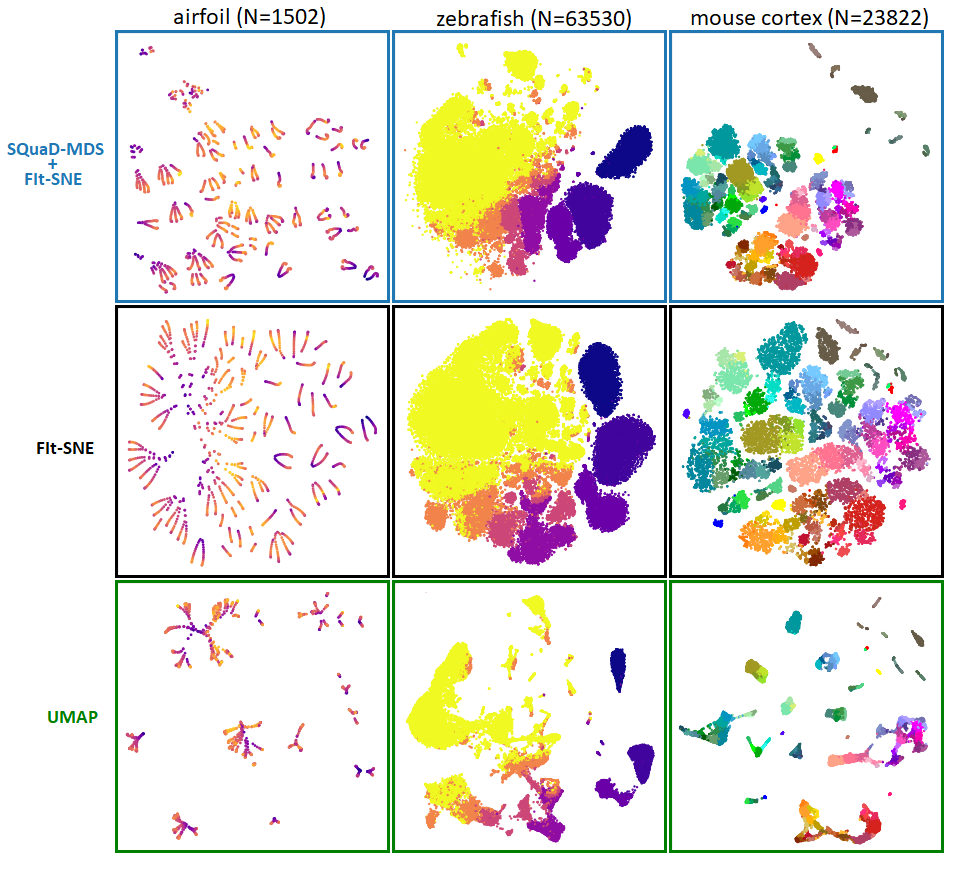}
        \caption{Embeddings produced by the proposed hybrid method, FI$t$-SNE, and UMAP.}
        \label{fig:embeddings_hybrid}
    \end{figure}
\end{center}

A qualitative comparison of the methods can be observed in Fig.~\ref{fig:embeddings_hybrid}. Visually, the global organisation of the points in the embeddings produced by the hybrid method is close to the organisation observed in the mMDS embeddings of Fig.~\ref{fig:MDSembeddings}. However, the points tend to form the sharp and easy-to-differentiate clusters that are typical of neighbour embeddings \cite{vanDerMaaten2008}. 

In the zebra fish data set, the colours correspond to an amount of time since fertilisation of the embryo. On the bottom of the embedding produced by the hybrid method, the transition from purple to yellow representing the passing time is more linear from right to left than in the FI$t$-SNE embedding, revealing another point of view on the organisation of the data. 

Similarly, the hierarchical organisation of the mouse cortex data set is clearly visible in the embedding produced by the hybrid method. The gray non-neuron cells are kept far from the coloured neurons, and the neurons are themselves organised in two large clusters of inhibitory (warm colours) and excitatory (cold colours) neurons; some finer-grained structures are also apparent within each cluster. 

\section{Conclusion}
\label{sec:concl}

The fast distance-preserving gradients proposed in this work enable the application of mMDS to large data sets. Empirically, the large-scale and medium-scale structures of the data are preserved similarly between SQuaD-MDS and the popular SMACOF algorithm, but the small-scale structures can suffer from the stochastic nature of SQuaD-MDS on some data sets. It is however noteworthy that the quadratic time complexity of the SMACOF algorithm with respect to the number of data points prohibits its application on large data sets, while the linear time complexity of SQuaD-MDS is considerably more affordable. 

Experiments show that merging the SQuaD-MDS gradients with those of a neighbour embedding method can yield embeddings that both preserve the large-scale structures of the data and the local structures, thereby addressing an often reported issue of almost exclusively local methods like $t$-SNE and its variants, including UMAP. 

Multiple perspectives arise from the quartet-based computation of distance gradients. For instance, the HD distances computed on each considered quartet could be first transformed nonlinearly to emphasize either large or small distances, or even to mitigate the effects of the concentration of norms, which tends to make the application of MDS challenging on very high dimensional data sets \cite{concNorms}. Yet another possibility is to consider a different strategy when sampling the quartets; for instance, increasing the probability to regroup points that are close by in the embedding when defining the quartets could help to better preserve the local structures of data. 
%for instance bringing a higher probability to regroup points into a quartet when they are close by in the embedding could help preserve the local structures of data.

\section{Acknowledgements}

This work was supported by Service Public de WallonieRecherche under grant n° 2010235 -ARIAC by DIGITALWALLONIA4.AI. JAL is a Senior Research Associate with the Fonds de la Recherche
Scientifique-FNRS.

\section*{} % references

\bibliography{mybibfile}

\end{document}